# Interpretable Feature Learning Framework for Smoking Behavior Detection

Nakayiza Hellen[1] and Ggaliwango Marvin[2]

[1,2] Brac University, Dhaka, Bangladesh
nakayiza.hellen@g.bracu.ac.bd, ggaliwango.marvin@g.bracu.ac.bd

**Abstract.** Smoking in public has been proven to be more harmful to non-smokers, making it a huge public health concern with urgent need for proactive measures and attention by authorities. With the world moving towards the 4th Industrial Revolution, there is a need for reliable eco-friendly detective measures towards this harmful intoxicating behavior to public health in and out of smart cities.

We developed an Interpretable feature learning framework for smoking behavior detection which utilizes a Deep Learning VGG-16 pretrained network to predict and classify the input Image class and a Layer-wise Relevance Propagation (LRP) to explain the network detection or prediction of smoking behavior based on the most relevant learned features/ pixels or neurons. The network's classification decision is based mainly on features located at the mouth especially the smoke seems to be of high importance to the network's decision. The outline of the 'smoke' is highlighted as evidence for the corresponding class. Some elements are seen as having a negative effect on the neuron "smoke" and are consequently highlighted differently. It is interesting to see that the network distinguishes important from unimportant features based on the image regions. The technology can also detect other smokeable drugs like weed, shisha, marijuana etc. The framework allows for reliable identification of action-based smokers in unsafe zones like schools, shopping malls, bus stops, railway compartments or other 'violated' places for smoking as per the government's regulatory health policies. With installation clearly defined in smoking zones, this technology can detect smokers out of range.

## 1 Introduction

### 1.1 Background and Motivation

Public health surveillance in smart cities require reliable technological approaches to monitor healthy life styles of populations in an eco-friendly manner. Unfortunately, smoking is on lifestyle that not only intoxifies to the environments but also harms the non-smokers if done in public. Being a great contributor to lung cancer, heart disease, maternal, morbidity and bronchitis, the smoking behavior is increasing overall health care costs and environmental pollution in smart cities.

The 4th Industrial Revolution and smart city trends necessitate trustable technological approaches to smoking behavior detection in public no matter the style used or type of material being smoked thus cigarettes weed, shisha, marijuana etc. Existing technologies unreliably detect violators of smoking policies based on image sequencing and smoke detectors and are expensively monitored and maintained for accurate detection results. They ignore the various alterations in smoking styles, patterns and behavior the process of detection is ambiguous (not explainable).

The challenges above and advancement machine vision is what motivated to proposing, develop and test an Interpretable Feature Learning for Smoking Behavior detection solution with explainable and trustable detections to improve public health monitoring and surveillance in smart cities for a healthier environment.

### 1.2 Research Contributions

The main contributions of this study include:

- We propose, develop and test a feature driven explainable approach to smoking behavior detection using the VGG-16 Network
- We propose and deploy detection interpretability with LRP for detection understandability of the learned smoking features
- We also give a visual comparison interpretability of the deep learned smoking behavioral explainability provided by the VGG-16 Network.

### 1.3 Research Contributions

The rest of the paper is arranged as follows: Section 2 includes Existing Works, Section 3 deals with the methodology, and section 4 contains Results and Discussion. The Concluding remarks and Future works are mentioned in section 5.

## 2 Existing works

### 2.1 Smoking Behavior

Smoking behavior analysis is all about finding related temporal patterns. Seasonality of smoking may have major implications, onset of youth smoking, the initiation of smoking among adolescents [3][4][5][6][7]

Smokers' smoking behaviors are affected by many factors and the randomness is very high which can be easy to be affected by the real environment. Therefore there is need to find more effective ways to analyze smoking behaviors and find new ways to quit smoking, and accurate prediction of People's Daily smoking time[8].

### 2.2 Smoking Behavior Detection

Accurately monitoring and modeling smoking behavior in real life settings is critical for designing and delivering appropriate smoking-cessation interventions through mobile Health applications [9].Video cameras have been used to capture movie frames of smoking behavior onto which image processing and analysis algorithms have been applied to form a time-series smoking topography that quantified various smoking behavior parameters [10].

### 2.3 Feature Learning

Feature learning focuses on abstract features with semantic information instead of concrete ones with details. Many existing unsupervised feature learning algorithms that learn these features from the data itself involve an unsupervised training stage where the algorithm learns a parameterized feature representation from the training data. Several Neural networks have been used to perform feature learning, since they learn a representation of their input at the hidden layer(s) which is subsequently used for classification or regression at the output layer [11]. In a comparative evaluation of unsupervised feature learning methods, Coates, Lee and Ng found that RBMs outperforms the rest on an image classification task [12]. The main demerit of many feature learning systems is their complexity. A requirement for such algorithms is careful selection of multiple hyper parameters for instance learning rates, momentum, sparsity penalties, weight decay, and so on must be chosen through cross-validation [13].

### 2.4 VGG-16 Network

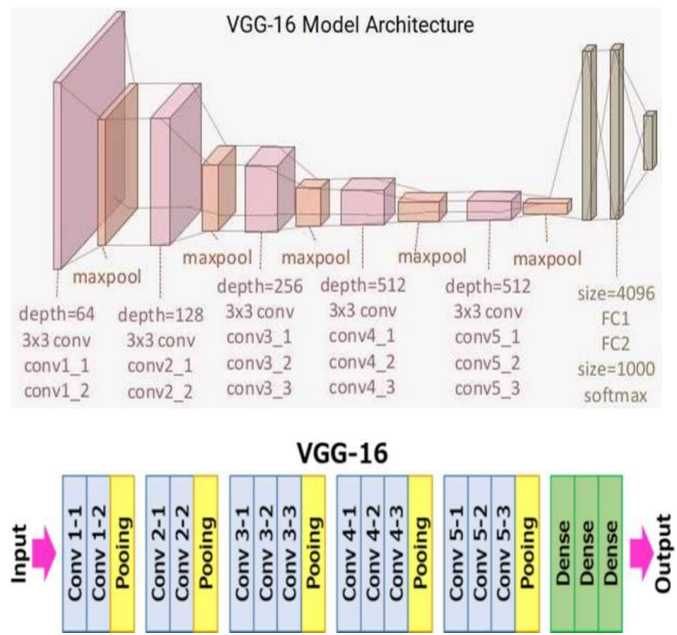

**Fig. 1.** VGG-16 Network.

An RGB image of a fixed size 224x224 is input to the conv1 layer, passed through a stack of convolutional (conv.) layers, which use filters with a very small receptive field: 3×3. It can also utilize 1×1 convolution filters as a linear transformation of the input channels (followed by non-linearity) in another configuration. The convolution stride is fixed to 1 pixel that is to say the spatial padding is 1-pixel for 3×3 conv. layers.

Spatial pooling is done by five max-pooling layers, which follow some of the conv. Layers. Max-pooling is done over a 2×2 pixel window, with stride 2.

A stack of convolutional layers is followed by three fullyconnected(FC) layers whose configuration is the same in all networks: the first two have 4096 channels each, the third performs 1000-way ILSVRC classification and thus contains 1000 channels (one for each class). Softmax is the final layer.

All hidden layers are equipped with the rectification (ReLU) non-linearity. Except for one, none of these networks contains Local Response Normalization (LRN) which increases memory consumption and computation time instead of performance improvement on the ILSVRC dataset.

Disadvantageously, VGG-Net is really very slow to train with quite large network architecture weights (concerning disk/bandwidth).VGG16 is over 533MB (its depth and number of fully-connected nodes) which makes its deployment a tiresome task and it is used in many deep learning image classification problems.

## 2.5     Existing Explainability Methods for Image detection/ Classification

**Local Interpretable Model-Agnostic Explanations (LIME).**
The LIME algorithm illuminates black box predictions of any classifier (f) in a correct way, by estimating it locally with an interpretable model g\in G$g \in G$. where GG is a class of interpretable models such as a linear classifier or a decision tree. The measure of complexity \Omega (g)$\Omega(g)$ of the model is also a significant factor of how easily the explanations are generated. In addition, the error of gg is calculated in approximating ff using a loss or distance function, denoted as L(f,g)$L(f,g)$. Finally the explanation \xi (g)$\xi(g)$ is calculated from the optimization of; $\xi(g)\Omega = \text{argmin} L(f,g) + \Omega(g)$

LIME is used to train a linear model to approximate the local decision boundary for that instance in the dataset, its used by non-experts to pick classifiers that generalize better in the real world and improve trustworthiness of classifiers by doing feature engineering with guidance on when and why to trust a model. With respect to a single prediction, this model-agnostic method generates an explanation by training a local interpretable classifier where its training data is generated by taking a specific input, permuting it, and labeling the permutations using the model [12][13][14].

In LIME, the predictions of the data set are then used as labels to train an interpretable linear model. The samples are weighted according to the proximity to the instance of relevance [15].

**SHapley Additive exPlanations (SHAP).** In 2017, Lundberg and Lee published a game theoretical approach that explains outputs of ML models by connecting optimal credit allocation with local explanations using the Shapley values of game theory and their related extensions which created the SHAP AI framework. This average marginal contribution of a feature value over all possible coalitions defines the Shapley value which are unified measures of a feature importance derived by; where the marginal contribution of the feature [p(S U i) – p(S)] is calculated over all the subsets S to obtain the Shapley value for a feature i, thus the model estimate is computed for all the subsets with and without the feature and summed to get the Shapley value for that feature [14] In SHAP[15], Shapley values of the segments are approximated, which reflect their contribution to the prediction.

**Explain it Like I'm 5 (ELI5).** ELI5 is a python package with in-built support for several ML frameworks to inspect ML classifiers and explain their predictions. It allows for visualization and debugging of various machine learning algorithms such as sklearn regressors and classifiers, Boost, CatBoost, Keras etc.using a unified API. By providing weights of the features from most common Python Libraries and has local (how and why a specific prediction is made) and global (how an overall model works) properties, ELI5 explains the ML models interpretably and computes the contribution selected features to execute feature importance for final prediction [17][18].

**Gradient-weighted Class activation Mapping (Grad-CAM).** Grad-CAM [19] is an extended work based on CAM (where the GAP layer before the output layer is not necessary); uses the gradients with respect to the target class c that flows to the final convolutional layer. GradCAM produces a coarse localization map of width v and height u, which highlights the important pixels for the classification of the image. The gradient of the class score is calculated with respect to the activation maps of the last convolutional layer. The gradients flow back after being averaged over the activation map's size Z and then the neuron's importance weights are calculated as:

$$L^c_{\mathbf{Grad-CAM}} \in \mathbb{R}^{v \times u}$$

$$\frac{\partial y^c}{\partial \mathbf{A_k}}$$

$$a^c_k = \frac{1}{Z} \sum_i \sum_j \frac{\partial y^c}{\partial A_k(i,j)}$$

$$\alpha^c_k$$

The weighting factor shows the importance of feature k for class c. Finally, the GradCAM heat maps are produced using the forward propagation activations as:

$$L^c_{\mathbf{Grad-CAM}} = ReLU(\sum_k a^c_k \mathbf{A_k})$$

In [20] the authors employed Grad-CAM (Selvaraju et al., 2017) to highlight the most important features of the input

**Search for Evidence Counterfactual (SEDC) for Image Classification.** In electronic format, an image comprises pixel values (one value per pixel for grayscale images, three values (RGB) per pixel for colored images) and these individual pixel values are typically used as input features for an image classifier. With interpretable concepts in images being embodied by groups of pixels or segments, authors of [22] performed a segmentation, similarly to LIME [21] and SHAP [16]. A label was assigned to each individual pixel reflecting the segment the pixel belongs to. The intention is to find a small set of segments that would, in case of not being present, alter the image classification. SEDC [18] was applied to binary document classification tasks with a single prediction score reflecting the probability of belonging to the class of interest as output. There are usually more than two possible categories, each with its own prediction score in image classification applications thus SEDC is generalized by enabling the occurrence of multiclass problems. Precisely, additional segments are selected by looking for the highest reduction in predicted class score. Consider an image I assigned to class c by a classifier CM. Like in [22], the aim is to find a counterfactual explanation E of the following form: an irreducible set of segments that leads to another classification after removal. This description can be formalized as:

$$E \subseteq I \text{ (segments in image)} \tag{1}$$

$$C_M(I \setminus E) \neq c \text{ (class change)} \tag{2}$$

$$\forall E' \subset E : C_M(I \setminus E') = c \text{ (irreducible)} \tag{3}$$

The counterfactual explanation can be complemented with the image causing the class change after removing the segments and the corresponding class.

**Layer-Wise Relevance Propagation (LRP).** Bach et al. [23] introduced Layer-Wise Relevance Propagation (LRP) as a model-specific method to create instance-level explanations for neural networks.[24] It operates by building for each neuron of a deep network a local redistribution rule, and applying these rules in a backward pass in order to produce the pixel-wise decomposition. The significance of individual input pixels for a certain class is calculated by redistributing the prediction scores backwards through the network. LRP [25] depends on decomposition of the decision and produces relevance scores between the activations $x(i)$ of neuron $i$ and its input, and finds the neuron's importance scores $R^{l}(i)$ at the layer $l$. More specifically, the relevance scores $R^{l}(i)$ of layer $l$ are calculated with respect to the layer $l+1$ as :

$$R^l(i) = \sum_j \frac{x(i)w(i,j)}{\sum_i x(i)w(i,j)} R^{l+1}(j)$$

where $w(i,j)$ is the weight between neuron $i$ and neuron $j$ [24] LRP has been successfully applied to many different models and tasks beyond classification of images by convolutional neural networks. It was applied to Bag-ofWords models in [26] and [13] to Fisher Vector / SVM classifiers respectively. In [14] it was used for the identification of relevant words in text documents and for visualizing facial features related to age, happiness and attractivity [15]. Also the authors of [16] used LRP for identifying relevant spatio-temporal EEG features in the context of Brain-Computer Interfacing. In [27] layer-wise relevance propagation (LRP), specifically -LRP (Bach et al., 2015b), and DeepLIFT (Shrikumar et al., 2017) were used to assign relevance scores to every word in the input text. Words with the highest and the lowest scores were selected as evidence for and counter-evidence against the predicted class, respectively.

LRP method was adapted to decompose the predictions of a convolutional neural network (CNN) and a bag-of-words SVM classifier models onto words on a topic categorization task [28]. Resulting scores showed how much individual words contribute to the overall classification decision and they further used the word-wise relevance scores to generate novel vector-based document representations which capture semantic information.

## 3 Methodology

### 3.1 Proposed Framework

The need for public health besides advancement of deep learning for machine vision and rapid spread of cameras in smart cites aided our proposal for a computer vision solution to smoke behavior detection using a VGG16 pertained deep neural network for feature learning and classification.

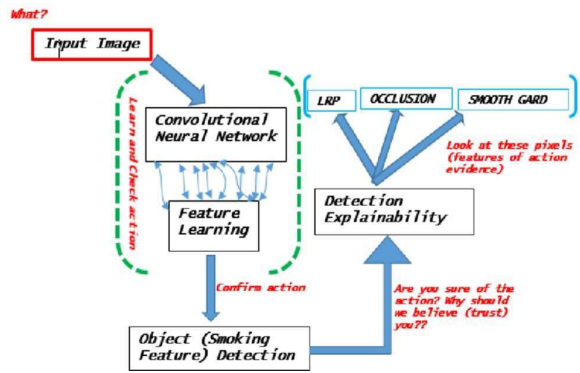

**Fig. 2.** Proposed Framework for Smoking behavior detection.

This explainable neural network is evaluated according to its performance with its detection decisions explained using Layer-wise Relevance Propagation (LRP), Occlusion analysis, and Integrated Gradient (SmoothGrad) whose interpretability of learned features is compared to assess the trustworthiness of the smoke behavior detection based on the most relevant smoking features learned during training.

### 3.2    Dataset Description

In this research, we have used the SmokerVsNonsmoker Dataset [29] containing smoking and not-smoking images. The dataset contains a total of 2400 raw images, where 1200 images are of smoking (smokers) category and remaining half belongs to no smoking (non-smokers) category. For a certain degree of inter-class confusion (for better model training), versatile images in both classes are considered for instance;
•  The smoking category comprises images of smokers from multiple angles and various gestures.
•  The not-smoking category contains images of nonsmokers with slightly similar gestures as that of smoking images such as people drinking water, using inhaler, holding the mobile phone, biting nails etc.

We have also used a collection of images from Google searches using several keywords such as cigarette smoking, smoker, person, coughing, taking inhaler, person on the phone, drinking water etc.) From the results of the query, it produces several suggestions to some relevant images. Even though each class consists of 100 original images, only images with high resolution above 1,000 pixels are selected to get good quality images.

### 3.3    Data preparation and Processing

We select six images randomly and scaled them to obtain similar sizes. We then perform general data visualization to recognize and learn any features that are indicative of smoking. In this implementation, we focus on only two classes (smoking/non-smoking) and save the respective image labels.

### 3.4    Implementation of the VGG-16 Network

Here, we take the VGG-16 pretrained network for image classification. For this network, we consider the task of explaining the evidence for the class "smoking" or "nonsmoking" it has found in the input image. The image is first loaded in the notebook and is then converted to a torch tensor of appropriate dimensions and normalized to be given as input to the VGG-16 network. The VGG-16 network is then loaded and its top-level dense layers are converted into equivalent 1x1 convolutions. The input can then be propagated in the network and the activations at each layer are collected. Activations in the top layer are the scores the neural network predicts for each class. We observe that the neuron 'smoking' (index 1066) has the highest score. This is expected due to the presence of 'smoke' in the image. Other related classes are also assigned a high score, as well as classes corresponding to other objects present in the image (e.g. smoke pipe, cigarette, lighter etc.).

### 3.5    Explainability with LRP

We now apply the layer-wise relevance propagation (LRP) procedure from top to bottom layers of the network in reverse order and apply propagation rules at each layer. As the first step, we create a list to store relevance scores at each layer. Top layer relevance scores are set to the toplayer activations and multiplied by a label indicator (the mask in order to retain only the evidence for the actual class "smoking". This evidence can then be propagated backward in the network by applying propagation rules at each layer, applying this sequence of computations. Fig. 2. Proposed Framework for Smoking behavior detection.

Convolution layers: Knowing that convolutions are special types of linear layers, we use the propagation rules, and a four-step procedure for applying these rules. The seconds and fourth steps are simple element-wise computations. The first step can be executed as a forward computation in the layer (where we have preliminary transformed the layer parameters and applied the increment function afterwards). The third step can instead be computed as a gradient in the space of input activations.

Pooling layers: max computed as a gradient in the space of input activations:- pooling layers are treated as average pooling layers in the backward pass. Knowing that average pooling is also a special linear layer, the same propagation rules as for the convolutional layers become applicable. In the code, the propagation procedure is iterated from the top-layer towards the lower layers. Every max-pooling layer is converted into an average pooling layer. Relevance scores at each layer can be visualized as a two-dimensional maps since each layer comprises a collection of two-dimensional feature maps.

In this case, relevance scores are pooled over all feature maps at a given layer and the 2D maps are shown for a selection of VGG-16 layers.

We observe that the explanation becomes increasingly resolved spatially and the loop above stops one layer before reaching the pixels. For propagation of relevance scores until the pixels, zB-rule (that properly handles pixel values received as input)is applied to this last layer; given by:

$$R_i = \sum_j \frac{a_i w_{ij} - l_i w_{ij}^+ - h_i w_{ij}^-}{\sum_i a_i w_{ij} - l_i w_{ij}^+ - h_i w_{ij}^-} R_j$$

This same rule can again be applied in terms of forward passes and gradient computations. We have now reached the bottom layer. The obtained pixel-wise relevance scores can now be summed over the RGB channels to indicate actual pixel-wise contributions and be rendered as a heatmap.

We observe that the heatmap highlights the outline of the 'smoke' as evidence for the corresponding class. Some elements are seen as having a negative effect on the neuron "smoke" and are consequently highlighted in blue. Simply put, relevant pixels are highlighted in Red and pixels that contribute negatively to the prediction, if any, are shown in blue.

# 4    Results and Discussion

## 4.1    Performance Evaluation

**Table 1.** Classification Report.

| Metric | Precision | Recall | F1-Score | Support |
|---|---|---|---|---|
| Non-smoking | 0.89 | 0.99 | 0.94 | 805 |
| Smoking | 0.98 | 0.88 | 0.93 | 805 |
| Accuracy | | | 0.93 | 1610 |
| Macro Average | 0.94 | 0.93 | 0.93 | 1610 |
| Weighted Average | 0.94 | 0.93 | 0.93 | 1610 |
| Accuracy Score: 0.93 | | | | |

**Table 2.** Testing Results

| Metric | Value |
|---|---|
| Test Loss | 0.19 |
| Test Accuracy | 93.35% |
| Test AUC | 0.99 |
| Test Precision | 0.98 |
| Test Recall | 0.88 |

After observing Table 1 and Table 2, we can conclude that a large number of epoch provides better accuracy for the training and testing phases. However, large number of epochs takes longer time to produce output.

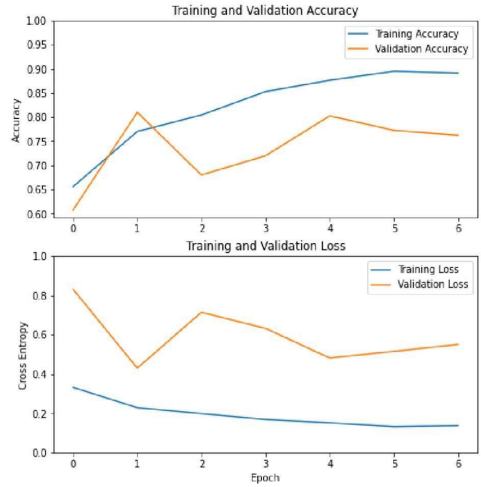

**Fig. 3.** Training and Validation Accuracy and Loss.

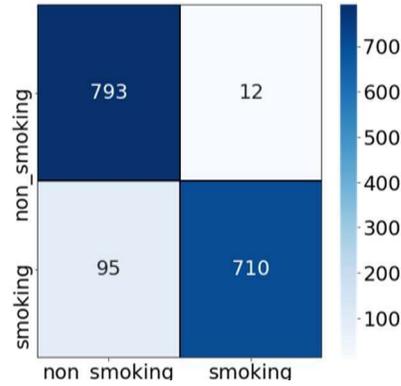

**Fig. 4.** Confusion Matrix

Fig.3 illustrates the results for accuracy and loss during Training and Validation and Fig. 4 shows the computed confusion matrix.

### 4.2 Simulation of Results for Explainability Methods

Looking at the computed relevance maps (Fig. 5 and Fig. 6), it turns out that the network's classification decision is based mainly on features located at the mouth especially the smoke seems to be of high importance to the network's decision. It is interesting to see that image regions with high contrasts are not subject to high relevance scores, indicating that the network has learned to distinguish important from unimportant features.

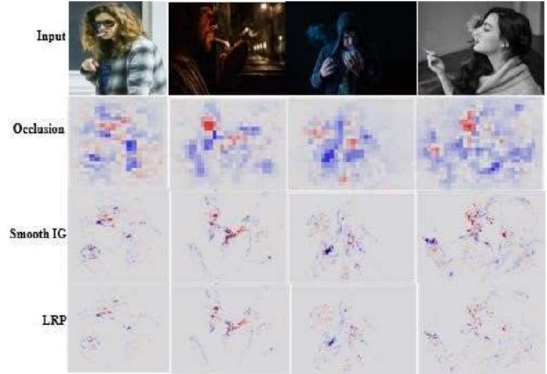

**Fig. 5.** Relevance Maps for Smokers.

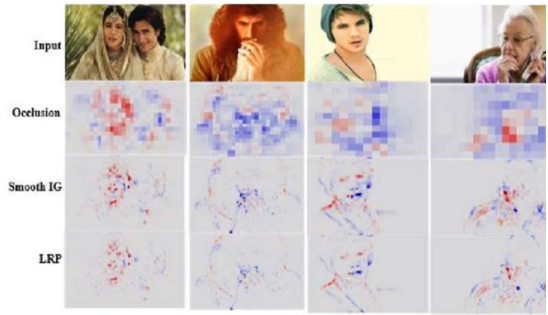

**Fig. 6.** Relevance Maps for Non-Smokers.

## 5    Conclusion and Future works

In this paper, we try to improve smoking behavior detection in unsafe zones of smart cities. To achieve this motivation, we proposed a smoking behavior detection framework. In the proposed framework, we make full use of the VGG-16 Network to predicts and classify the class/category of the input Image. Also we proposed the LRP technique to explain the network prediction base on the most relevant pixels or neurons. To evaluate our proposed framework, we provide a classification report and illustrate values for the Loss function and Accuracy score of our framework throughout training and validation. As far as smoking behavior detection is concerned, this approach can be implemented in public places like a campus, school, shopping malls, bus stops, railway compartments or any other 'violated' places for smoking as per a country's government regulations. In future, we hope to improve our proposed framework by using real data collection and modifying the framework to include external factors related to smoking behavior.